\documentclass[leqno]{article}
\usepackage[logonly]{trace}

\usepackage{hltnaacl04}
\usepackage{amsmath}
\usepackage{amssymb}
\usepackage{txfonts}
\usepackage{courierc}
\usepackage{url}
\usepackage{prooftree}
\usepackage{comment}
\usepackage{hyphenat}
\usepackage{semantics}
\usepackage{exemplar}
\usepackage{refrange}
\usepackage{varioref}
\usepackage{fixltx2e}
\usepackage{dblfloatfix}
\usepackage{mdwtab}
\usepackage{wrapfig}
\usepackage[matrix,arrow,curve,frame]{xy}

\usepackage{natbib1}
\makeatletter
\def\citename#1#2(@)(@)\@nil#3{\expandafter\NAT@apalk#1, #2, \@nil{#3}}
\makeatother

\divide\floatsep 2
\divide\dblfloatsep 2
\divide\intextsep 2
\makeatletter
\@maxdepth\dp\strutbox \global\maxdepth\@maxdepth
\makeatother

\newcommand{\fuse}[1][]{\circ_{#1}}
\newcommand{\into}[1][]{\backslash_{#1}}
\newcommand{\from}[1][]{/_{#1}}
\newcommand{\Boxd}{\Box^{\scriptscriptstyle\downarrow}}

\newcommand{\s}{\Diamond_u\mathit{s}}
\newcommand{\sPos}{\Diamond_u\Boxd_p\Diamond_p\mathit{s}}
\newcommand{\sNeg}{\Boxd_p\Diamond_p\Diamond_u\mathit{s}}
\newcommand{\np}{\mathit{np}}

\newcommand{\Left}{\mathrm{Left}}
\newcommand{\Right}{\mathrm{Right}}
\newcommand{\Root}{\mathrm{Root}}
\newcommand{\Axiom}{\mathrm{Axiom}}
\newcommand{\Intro}{\mathrm{I}}
\newcommand{\Elim}{\mathrm{E}}
\newcommand{\Quote}{\mathrm{T}}
\newcommand{\Concat}{\mathrm{K}'}
\newcommand{\Unquote}{\mathrm{Unquote}}

\makeatletter
\renewcommand{\fuse}[1][]{\circ\ifx\@let@token[_{#1}\fi}
\renewcommand{\from}[1][]{/\ifx\@let@token[_{\mskip-1.5\thinmuskip#1}\fi}
\makeatother

\begin{document}
\title{Polarity sensitivity and evaluation order in type-logical grammar}
\author{Chung-chieh Shan\\
    Division of Engineering and Applied Sciences\\
    Harvard University\\
    Cambridge, MA 02138\\
    \url{ccshan@post.harvard.edu}}
\maketitle

\begin{abstract}
We present a novel, type-logical analysis of \emph{polarity sensitivity}: how
negative polarity items (like \phrase{any} and \phrase{ever}) or positive ones
(like \phrase{some}) are licensed or prohibited.  It takes not just scopal
relations but also linear order into account, using the programming\hyp language
notions of delimited continuations and evaluation order, respectively.  It thus
achieves greater empirical coverage than previous proposals.
\end{abstract}

\multiply\abovedisplayskip 2
\multiply\belowdisplayskip 2
\multiply\abovedisplayshortskip 2
\multiply\belowdisplayshortskip 2
\divide\abovedisplayskip 5
\divide\belowdisplayskip 5
\divide\abovedisplayshortskip 5
\divide\belowdisplayshortskip 5

\section{Introduction}
\label{s:introduction}

Polarity sensitivity \citep{ladusaw-polarity} has been a popular
linguistic phenomenon to analyze in the categorial \citep{dowty-role},
lexical\hyp functional \citep{fry-negative,fry-proof}, and type\hyp logical
\citep{bernardi-reasoning,bernardi-generalized} approaches to grammar.  The
multitude of these analyses is in part due to the more explicit emphasis that
these traditions place on the syntax-semantics interface---be it in the form of
Montague-style semantic rules, the Curry-Howard isomorphism, or linear
logic as glue---and the fact that polarity sensitivity is a phenomenon
that spans syntax and semantics.

On one hand, which polarity items are licensed or prohibited in a given
linguistic environment depends, by and large, on semantic properties of that
environment \citep[inter alia]{ladusaw-polarity,krifka-semantics}.  For
example, to a first approximation, negative polarity items can occur only in
\emph{downward\hyp entailing} contexts, such as under the scope of
a \emph{monotonically decreasing} quantifier.
A quantifier~$q$, of type $(e\toF
t)\toF t$ where $e$ is the type of individuals and $t$ is the type of truth
values, is monotonically decreasing just in case
\begin{equation}
    \Forall{s_1} \Forall{s_2}
    \bigl(\Forall{x} s_2(x) \limplies s_1(x)\bigr)
        \limplies q(s_1) \limplies q(s_2).
\end{equation}
Thus \eqref{e:no-any} is acceptable because the scope of \phrase{nobody} is
downward\hyp entailing, whereas (\refrange{e:every-any}{e:alice-any}) are
unacceptable.
\example{
\subexamples{\label{e:semantic}%
\subexample{\label{e:no-any}Nobody saw anybody.}
\subexample{\<*\label{e:every-any}Everybody saw anybody.}
\subexample{\<*\label{e:alice-any}Alice saw anybody.}}}

On the other hand, a restriction on surface syntactic form, such as that imposed
by polarity sensitivity, is by definition a matter of syntax.  Besides, there
are syntactic restrictions on the configuration relating the licensor to the
licensee.  For example, \eqref{e:no-any} above is acceptable---\phrase{nobody}
manages to license \phrase{anybody}---but \eqref{e:any-no} below is not.  As
the contrast in~\eqref{e:deeper-linear} further illustrates, the licensor
usually needs to precede the licensee.
\examples{
\example{\label{e:any-no}\<*Anybody saw nobody.}
\example{\label{e:deeper-linear}%
\subexamples{
\subexample{\label{e:licensing-out}Nobody's mother saw anybody's father.}
\subexample{\<*Anybody's mother saw nobody's father.}}}}
The syntactic relations allowed between licensor and licensee for polarity
sensitivity purposes are similar to those allowed between antecedent and
pronoun for variable binding purposes.  To take one example, just as an
antecedent's ability to bind a (c\nobreakdash-commanded) pronoun percolates up
to a containing phrase (such as in~\eqref{e:binding-out}, what
\citet{buring-situation} calls ``binding out of DP''), a licensor's ability to
license a (c\nobreakdash-commanded) polarity item percolates up to a containing
phrase (such as in~\eqref{e:licensing-out}).
\example{\label{e:binding-out}[Every boy$_i$'s mother] saw his$_i$ father.}
Moreover, just as a bindee can precede a binder in a sentence when the bindee
sits in a clause that excludes the binder (as in~\eqref{e:binding-back}; see
\citealp{williams-blocking}, \S2.1), a licensee can precede a licensor in
a sentence when the licensee sits in a clause that excludes the licensor (as
in~\eqref{e:licensing-back}; see \citealp{ladusaw-polarity}, page~112).
\examples{
\example{\label{e:binding-back}That he$_i$ would be arrested for speeding came as a surprise to every$_i$ motorist.}
\example{\label{e:licensing-back}That anybody would be arrested for speeding came as a surprise to the district attorney.}}


This paper presents a new, type-logical account of polarity sensitivity that
encompasses the semantic properties exemplified in~\eqref{e:semantic} and the
syntactic properties exemplified in~(\refrange{e:any-no}{e:deeper-linear}).
Taking advantage of the Curry-Howard isomorphism, it is the first account
of polarity sensitivity in the grammatical frameworks mentioned above to
correctly fault~\eqref{e:any-no} for the failure of \phrase{nobody} to appear
before \phrase{anybody}.  The analysis makes further correct predictions,
as we will see at the end of~\S\ref{s:polarity}.

The analysis here borrows the concepts of \emph{delimited continuations}
\citep{felleisen-theory,danvy-abstracting}
and \emph{evaluation order} from the study of programming languages.  Thus this
paper is about computational linguistics, in the sense of applying insights from
computer science to linguistics.  The basic idea transfers to other formalisms,
but type-logical grammar---more precisely, multimodal categorial
grammar---offers a fragment~$\mathrm{NL}\Diamond_{\mathcal{R}^-}$ whose parsing
problem is decidable using proof-net technology \citep[\S9.2]{moot-thesis},
which is of great help while developing and testing the theory.

\section{Delimited continuations}

Figure~\ref{fig:nd} shows natural deduction rules for multimodal
categorial grammar, a member of the type-logical family of grammar formalisms
\citep{moortgat-categorial,bernardi-reasoning}.
\begin{figure}
\vspace{-1ex}
\begin{gather*}
\begin{prooftree}
    \justifies A \vdash A \using \Axiom
\end{prooftree}
\end{gather*}

For each unary mode~$\alpha$ (blank, $u$, or~$p$ in this paper):
\begin{gather*}
\begin{prooftree}
    \Diamond_\alpha \Gamma \vdash A
    \justifies \Gamma \vdash \Boxd_\alpha A \using \Boxd_\alpha\Intro
\end{prooftree}
\qquad
\begin{prooftree}
    \Gamma \vdash \Boxd_\alpha A
    \justifies \Diamond_\alpha \Gamma \vdash A \using \Boxd_\alpha\Elim
\end{prooftree}
\\
\begin{prooftree}
    \Gamma \vdash A
    \justifies \Diamond_\alpha \Gamma \vdash \Diamond_\alpha A \using \Diamond_\alpha\Intro
\end{prooftree}
\qquad
\begin{prooftree}
    \Delta \vdash \Diamond_\alpha A \qquad
    \Gamma[\Diamond_\alpha A] \vdash B
    \justifies \Gamma[\Delta] \vdash B \using \Diamond_\alpha\Elim
\end{prooftree}
\end{gather*}

For each binary mode~$\beta$ (blank or~$c$ in this paper):
\begin{gather*}
\begin{prooftree}
    \Gamma \vdash B \quad \Delta \vdash C
    \justifies \Gamma \fuse[\beta] \Delta \vdash B \fuse[\beta] C \using \fuse[\beta] \Intro
\end{prooftree}
\qquad
\begin{prooftree}
    \Delta \vdash B \fuse C \hspace{.75em} \Gamma[B \fuse[\beta] C] \vdash A
    \justifies \Gamma[\Delta] \vdash A \using \fuse[\beta] \Elim
\end{prooftree}
\\
\begin{prooftree}
    \Delta \fuse[\beta] B \vdash C
    \justifies \Delta \vdash C\from[\beta]B \using \from[\beta]\Intro
\end{prooftree}
\qquad
\begin{prooftree}
    \Delta \vdash B\from[\beta]A \qquad
    \Gamma \vdash A
    \justifies \Delta \fuse[\beta] \Gamma \vdash B \using \from[\beta]\Elim
\end{prooftree}
\\
\begin{prooftree}
    B \fuse[\beta] \Delta \vdash C
    \justifies \Delta \vdash B\into[\beta]C \using \into[\beta]\Intro
\end{prooftree}
\qquad
\begin{prooftree}
    \Gamma \vdash A \qquad
    \Delta \vdash A\into[\beta]B
    \justifies \Gamma \fuse[\beta] \Delta \vdash B \using \into[\beta]\Elim
\end{prooftree}
\end{gather*}

\caption{Natural deduction rules for multimodal categorial grammar
    \citep[pages 9 and~50]{bernardi-reasoning}.  To reduce notation,
    we do not distinguish structural punctuation from logical
    connectives.}
\label{fig:nd}
\end{figure}
Figure~\ref{fig:postulates} lists our structural postulates.  These two figures
form the logic underlying our account.

\begin{figure}
\vspace{-1ex}
\divide\jot 2
\begin{align}
    \tag{$\Root$}   A &\dashv\vdash A \fuse[c] 1
\\  \tag{$\Left$}   (B \fuse C) \fuse[c] K &\dashv\vdash B \fuse[c] (C \fuse K)
\\  \tag{$\Right$}  (\Diamond B \fuse C) \fuse[c] K &\dashv\vdash C \fuse[c] (K \fuse \Diamond B)
\\
    \tag{$\Quote$}  A &\hphantom\dashv\vdash \Diamond A
\\  \tag{$\Concat$} \Diamond A \fuse \Diamond B &\hphantom\dashv\vdash \Diamond (A \fuse B)
\\  \tag{$\Unquote$}\Diamond \Diamond_u A &\hphantom\dashv\vdash \Diamond_u A
\end{align}
\caption{Structural postulates}
\label{fig:postulates}
\end{figure}

We use two binary modes: the default mode (blank) for surface syntactic
composition, and the continuation mode~$c$.  As usual, a formula of the form $A
\fuse B$ can be read as ``$A$ followed by~$B$''.  By contrast, a formula of the
form $A \fuse[c] B$ can be read as ``$A$ in the context~$B$''.  In
programming\hyp language terms, the formula $A \fuse[c] B$ plugs
a subexpression~$A$ into a delimited continuation~$B$.  The $\Root$ rule creates
a trivial continuation: it says that $1$ is a right identity for the $c$ mode,
where $1$ can be thought of as a nullary connective, effectively enabling
empty antecedents for the $c$ mode.  The binary modes, along with the first
three postulates in Figure~\ref{fig:postulates}, provide a new way to encode
Moortgat's ternary connective~$q$ \citeyearpar{moortgat-generalized} for
in-situ quantification.  For intuition, it may help to draw logical formulas as
binary trees, distinguishing graphically between the two modes.

To further capture the interaction between scope
inversion and polarity sensitivity exemplified
in~(\refrange{e:any-no}{e:deeper-linear}),
we use three unary modes: the value mode (blank), the unquotation mode~$u$, and
the polarity mode~$p$.  The value mode marks when an expression is devoid of
in-situ quantification, or, in programming\hyp language terms, when it is a pure
value rather than a computation with control effects.  As a special case, any
formula can be turned pure by embedding it under a diamond using
the $\Quote$ postulate, analogous to \emph{quotation} or \emph{staging} in
programming languages.  Quotations can be concatenated using the $\Concat$
postulate.  The unquotation mode~$u$ marks when a diamond can be canceled using
the $\Unquote$ postulate.  Unquotation is also known as \emph{eval} or
\emph{run} in programming languages.  The polarity mode~$p$, and the
empirical utility of these unary modes, are explained in~\S\ref{s:polarity}.

A derivation is considered complete if it culminates in a sequent whose
antecedent is built using the default binary mode~$\fuse$ only, and whose
conclusion is a type of the form $\Diamond_u A$.  Below is a derivation of
\phrase{Alice saw Bob}.
\begin{equation}
\begin{prooftree}
    \text{Alice} \vdash \np
    \[
        \text{saw} \vdash (\np\into\s)\from\np \quad
        \text{Bob} \vdash \np
        \justifies \text{saw} \fuse \text{Bob} \vdash \np\into\s
        \using \from\Elim
    \]
    \justifies \text{Alice} \fuse (\text{saw} \fuse \text{Bob}) \vdash \s
    \using \into\Elim
\end{prooftree}
\end{equation}
Note that clauses take the type~$\s$ rather than the usual~$\mathit{s}$,
so the $\Unquote$ rule can operate on clauses.  We abbreviate $\s$ to
\renewcommand{\s}{s^\circ}
$\s$ below.

\begin{figure*}
\begin{minipage}[b]{4.5in}
\vspace{-1ex}
\begin{gather*}
\begin{prooftree}
    \[ \[ \[ \[ \[
        \text{a man} \vdash \s\from[c](\np\into[c]\s)
        \hspace{-5em}
        \[
            \[ \[ \[ \[ \[ \[ \[ \[ \[
                \text{Alice} \vdash \np
                \[
                    \text{saw} \vdash (\np\into\s)\from\np
                    \[
                        \[ \justifies \np \vdash \np \using \Axiom \]
                        \text{'s mother} \vdash \np\into\np
                        \justifies \np\fuse\text{'s mother} \vdash \np
                        \using \into\Elim
                    \]
                    \justifies \text{saw}\fuse(\np\fuse\text{'s mother}) \vdash \np\into\s
                    \using \from\Elim
                \]
                \justifies \text{Alice}\fuse(\text{saw}\fuse(\np\fuse\text{'s mother})) \vdash \s
                \using \into\Elim \shiftright 2.5em
            \]
            \justifies \Diamond\bigl(\text{Alice}\fuse(\text{saw}\fuse(\np\fuse\text{'s mother}))\bigr) \vdash \Diamond\s
            \using \Diamond\Intro \shiftright 1em \]
            \justifies \Diamond\bigl(\text{Alice}\fuse(\text{saw}\fuse(\np\fuse\text{'s mother}))\bigr) \vdash \s
            \using \Unquote \]
            \justifies \Diamond\text{Alice}\fuse(\Diamond\text{saw}\fuse(\Diamond\np\fuse\Diamond\text{'s mother})) \vdash \s
            \using \Concat \text{ thrice} \shiftright .5em \]
            \justifies \Diamond\text{Alice}\fuse(\Diamond\text{saw}\fuse(\np\fuse\Diamond\text{'s mother})) \vdash \s
            \using \Quote \]
            \justifies \bigl(\Diamond\text{Alice}\fuse(\Diamond\text{saw}\fuse(\np\fuse\Diamond\text{'s mother}))\bigr)\fuse[c]1 \vdash \s
            \using \Root \shiftright 1em \]
            \justifies \bigl(\Diamond\text{saw}\fuse(\np\fuse\Diamond\text{'s mother})\bigr)\fuse[c](1\fuse\Diamond\text{Alice}) \vdash \s
            \using \Right \]
            \justifies (\np\fuse\Diamond\text{'s mother})\fuse[c]\bigl((1\fuse\Diamond\text{Alice})\fuse\Diamond\text{saw}\bigr) \vdash \s
            \using \Right \]
            \justifies \np\fuse[c]\bigl(\Diamond\text{'s mother}\fuse((1\fuse\Diamond\text{Alice})\fuse\Diamond\text{saw})\bigr) \vdash \s
            \using \Left \]
            \justifies \Diamond\text{'s mother}\fuse((1\fuse\Diamond\text{Alice})\fuse\Diamond\text{saw}) \vdash \np\into[c]\s
            \using \into[c]\Intro \shiftright .5em
        \]
        \justifies \text{a man}\fuse\bigl(\Diamond\text{'s mother}\fuse((1\fuse\Diamond\text{Alice})\fuse\Diamond\text{saw})\bigr) \vdash \s
        \using \from[c]\Elim
    \]
    \justifies (\text{a man}\fuse\Diamond\text{'s mother})\fuse[c]\bigl((1\fuse\Diamond\text{Alice})\fuse\Diamond\text{saw}\bigr) \vdash \s
    \using \Left \]
    \justifies \bigl(\Diamond\text{saw}\fuse(\text{a man}\fuse\Diamond\text{'s mother})\bigr)\fuse[c](1\fuse\Diamond\text{Alice}) \vdash \s
    \using \Right \]
    \justifies \bigl(\Diamond\text{Alice}\fuse(\Diamond\text{saw}\fuse(\text{a man}\fuse\Diamond\text{'s mother}))\bigr)\fuse[c]1 \vdash \s
    \using \Right \]
    \justifies \Diamond\text{Alice}\fuse(\Diamond\text{saw}\fuse(\text{a man}\fuse\Diamond\text{'s mother})) \vdash \s
    \using \Root \]
    \justifies \text{Alice}\fuse(\text{saw}\fuse(\text{a man}\fuse\text{'s mother})) \vdash \s
    \using \Quote \text{ thrice}
\end{prooftree}
\end{gather*}
\caption{In-situ quantification: deriving \phrase{Alice saw a man's mother}}
\label{fig:a-man}
\hbox{}
\end{minipage}%
\begin{minipage}[b]{2in}
\newcommand{\row}[3]{#1&#3\from[c](\np\into[c]#2)\\}
\renewcommand{\sPos}{s^+}
\renewcommand{\sNeg}{s^-}
\centering
\begin{tabular}{cMc}
\hlx{vhv}
    \textbf{Quantifier} & \text{\textbf{Type}}
\\
\hlx{vhv}
    \row{a man} \s \s
    \row{nobody} \sNeg \s
    \row{anybody} \sNeg \sNeg
    \row{somebody} \sPos \sPos
    \row{everybody} \sPos \s
\hlx{vhv}
\end{tabular}

\[\vcenter{\xymatrix @R=2.75em @C=2.75em{
        *=<2em,2em>[o][Fe]{\sPos}
            \ar[dr]_{\text{\small$\varepsilon$}} \ar@(ur,ul)+UR(.71);+UL(.71)_{\text{\small \smash[b]{somebody}}}
            \save !L *r@{>} \restore
    &&
        *=<2em,2em>[o][Fe]{\sNeg}
            \ar[dl]_{\text{\small$\varepsilon$}} \ar@(ur,ul)+UR(.71);+UL(.71)_{\text{\small \smash[b]{anybody}}}
    \\
    &
        *=<2em,2em>+[o][Fee]{\s}
            \ar@(r,d)[ur]_{\text{\small \hspace{-.5em}nobody}} \ar@(dl,dr)+DL(.71);+DR(.71)_{\text{\small a man}}
            \ar@(u,r)[ul]_{\text{\small \hspace{-.5em}everybody}}
            \save !L *r@{>} \restore
}}\]

\caption{Quantifier type assignments, and a corresponding finite-state machine}
\label{fig:polarities}
\hbox{}
\end{minipage}
\vskip-\baselineskip
\end{figure*}
To illustrate in-situ quantification, Figure~\vref{fig:a-man} shows a derivation
of \phrase{Alice saw a man's mother}.  For brevity, we treat \phrase{a man} as
a single lexical item.  It is a quantificational noun phrase whose polarity is
\emph{neutral} in a sense that contrasts with other quantifiers considered
below.  The crucial part of this derivation is the use of the structural
postulates $\Root$, $\Left$, and $\Right$ to divide the sentence into two parts:
the subexpression \phrase{a man} and its context \phrase{Alice saw \_'s mother}.
The type of \phrase{a man}, $\s\from[c](\np\into[c]\s)$, can be read as ``a
subexpression that produces a clause when placed in a context that can enclose
an $\np$ to make a clause''.

\section{Polarity sensitivity and evaluation~order}
\label{s:polarity}

The $p$ mode mediates polarity sensitivity.  For every unary mode~$\alpha$, we
can derive $A \vdash
\Boxd_\alpha\Diamond_\alpha A$ from the rules in Figure~\ref{fig:nd}.  This fact
is particularly useful when $\alpha = p$, because we assign the types $\sPos$
and $\sNeg$ to \emph{positive} and \emph{negative} clauses, respectively, and
can derive
\begin{align}
\label{e:subtyping}
\s&\vdash\sPos, & \s&\vdash\sNeg.
\end{align}
In words, a neutral clause can be silently converted into a positive or negative
one.  We henceforth write $s^+$ and~$s^-$ for $\sPos$ and $\sNeg$.
By~\eqref{e:subtyping}, both types are ``subtypes'' of~$\s$ (that is to say,
entailed by~$\s$).
\renewcommand{\sPos}{s^+}
\renewcommand{\sNeg}{s^-}

\begin{figure*}
\vspace{-2.25ex}
\begin{minipage}[b]{3.75in}
\begin{gather*}
\begin{prooftree}
    \[
        \text{nobody} \vdash \s\from[c](\np\into[c]\sNeg)
        \hspace{-8em}
        \[
            \[ \[ \[
            \text{anybody} \vdash \sNeg\from[c](\np\into[c]\sNeg)
            \hspace{-4em}
            \[ \[ \[ \[
                \[ \leadsto \Diamond\np\fuse(\Diamond\text{saw}\fuse\np) \vdash \s \]
                \justifies \np\fuse[c]\bigl((1\fuse\Diamond\np)\fuse\Diamond\text{saw}\bigr) \vdash \s
                \using \Root,\Right,\Right \]
                \justifies \Diamond_p\bigl(\np\fuse[c]((1\fuse\Diamond\np)\fuse\Diamond\text{saw})\bigr) \vdash \Diamond_p\s
                \using \Diamond_p\Intro \shiftright 2em \]
                \justifies \np\fuse[c]\bigl((1\fuse\Diamond\np)\fuse\Diamond\text{saw}\bigr) \vdash \sNeg
                \using \Boxd_p\Intro \shiftright 2em \]
                \justifies (1\fuse\Diamond\np)\fuse\Diamond\text{saw} \vdash \np\into[c]\sNeg
                \using \into[c]\Intro
            \]
            \justifies \text{anybody}\fuse[c]\bigl((1\fuse\Diamond\np)\fuse\Diamond\text{saw}\bigr) \vdash \sNeg
            \using \from[c]\Elim \]
            \justifies \Diamond\np\fuse[c]\bigl((\Diamond\text{saw}\fuse\text{anybody})\fuse1\bigr) \vdash \sNeg
            \using \Right,\Right,\Left \]
            \justifies \np\fuse[c]\bigl((\text{saw}\fuse\text{anybody})\fuse1\bigr) \vdash \sNeg
            \using \Quote \text{ twice} \shiftright 4em \]
            \justifies (\text{saw}\fuse\text{anybody})\fuse1 \vdash \np\into[c]\sNeg
            \using \into[c]\Intro
        \]
        \justifies \text{nobody}\fuse[c]\bigl((\text{saw}\fuse\text{anybody})\fuse1\bigr) \vdash \s
        \using \from[c]\Elim
    \]
    \justifies \text{nobody}\fuse(\text{saw}\fuse\text{anybody}) \vdash \s
    \using \Left,\Root
\end{prooftree}
\end{gather*}
\caption{Polarity licensing: deriving \phrase{Nobody saw anybody}}
\label{fig:no-any}
\hbox{}
\end{minipage}%
\begin{minipage}[b]{2.75in}
\begin{gather*}
\begin{prooftree}
    \[ \[ \[ \[ \[ \[
    \[ \leadsto \text{anybody}\fuse(\text{saw}\fuse\np) \vdash \sNeg \]
    \justifies \Diamond\bigl(\text{anybody}\fuse(\text{saw}\fuse\np)\bigr) \vdash \Diamond\sNeg
    \using \Diamond\Intro \]
    \leadsto \Diamond\bigl(\text{anybody}\fuse(\text{saw}\fuse\np)\bigr) \vdash \sNeg
    \using \text{???} \]
    \justifies \Diamond\text{anybody}\fuse(\Diamond\text{saw}\fuse\Diamond\np) \vdash \sNeg
    \using \Concat \text{ twice} \]
    \justifies \Diamond\text{anybody}\fuse(\Diamond\text{saw}\fuse\np) \vdash \sNeg
    \using \Quote \]
    \justifies \bigl(\Diamond\text{anybody}\fuse(\Diamond\text{saw}\fuse\np)\bigr)\fuse[c]1 \vdash \sNeg
    \using \Root \]
    \justifies (\Diamond\text{saw}\fuse\np)\fuse[c](1\fuse\Diamond\text{anybody}) \vdash \sNeg
    \using \Right \]
    \justifies \np\fuse[c]\bigl((1\fuse\Diamond\text{anybody})\fuse\Diamond\text{saw}\bigr) \vdash \sNeg
    \using \Right
\end{prooftree}
\end{gather*}
\caption{Linear order in polarity licensing: ruling out \phrase{Anybody saw nobody} using left-to-right evaluation order}
\label{fig:stuck}
\hbox{}
\end{minipage}
\vskip-\baselineskip
\end{figure*}
The $p$ mode is used in Figure~\vref{fig:no-any} to derive
\phrase{Nobody saw anybody}.  Unlike \phrase{a man}, the
quantifier \phrase{anybody} has the type $\sNeg\from[c](\np\into[c]\sNeg)$,
showing that it takes scope over a negative clause to make another negative
clause.  Meanwhile, the quantifier \phrase{nobody} has the type
$\s\from[c](\np\into[c]\sNeg)$, showing that it takes scope over a negative
clause to make a neutral clause.  Thus \phrase{nobody} can take scope over the
negative clause returned by \phrase{anybody} to make a neutral clause, which is
complete.

The contrast between \eqref{e:no-any} and~\eqref{e:any-no} boils down to
the $\Right$ (but not $\Left$) postulate's requirement that the leftmost
constituent be of the form~$\Diamond B$.  (In programming\hyp language terms,
a subexpression can be evaluated only if all other subexpressions to its left
are pure.)  For \phrase{nobody} to take scope over (and license)
\phrase{anybody} in~\eqref{e:any-no} requires the context
*\phrase{Anybody saw \_}.  In other words, the sequent
\begin{equation}
\label{e:stucker}
    \np\fuse[c]\bigl((1\fuse\Diamond\text{anybody})\fuse\Diamond\text{saw}\bigr) \vdash \sNeg
\end{equation}
must be derived, in which the $\Right$ rule forces the constituents \phrase{anybody} and
\phrase{saw} to be embedded under diamonds.  Figure~\ref{fig:stuck} shows an
attempt at deriving~\eqref{e:stucker}, which fails because the type $\sNeg$ for
negative clauses cannot be $\Unquote$d (shown with question marks).  The sequent
in~\eqref{e:stucker} cannot be derived, and the sentence *\phrase{Anybody saw
nobody} is not admitted.  Nevertheless, \phrase{Somebody saw everybody} is
correctly predicted to have ambiguous scope, because neutral and positive
clauses can be $\Unquote$d.

The quantifiers \phrase{a man}, \phrase{nobody}, and \phrase{anybody} in Figures
\ref{fig:a-man} and~\ref{fig:no-any} exemplify a general pattern of analysis:
every polarity\hyp sensitive item, be it traditionally considered a licensor or
a licensee, specifies in its type an \emph{input} polarity (of the clause it
takes scope over) and an \emph{output} polarity (of the clause it produces).
Figure~\ref{fig:polarities} lists more quantifiers and their
input and output polarities.  As shown there,
these type assignments can be visualized as a finite-state machine.
The states are the three clause types.  The $\varepsilon$\hyp
transitions are
the two derivability relations in \eqref{e:subtyping}.  The
non-$\varepsilon$ transitions are the quantifiers.  The start states are the clausal
types that can be $\Unquote$d.  The final state is the clausal type returned
by verbs, namely neutral.

The precise pattern of predictions made by this theory can be stated in two
parts.  First, due to the lexical types in Figure~\ref{fig:polarities} and the
``subtyping'' relations in~\eqref{e:subtyping}, the quantifiers in a sentence
must form a valid transition sequence, from widest to narrowest scope.  This
constraint is standard in type-logical accounts of polarity sensitivity.
Second, thanks to the unary modes in the structural postulates in
Figure~\ref{fig:postulates}, whenever two quantifiers take inverse rather than
linear scope with respect to each other, the transitions must pass through
a start state (that is, a clause type that can be $\Unquote$d) in between.  This
constraint is an empirical advance over previous accounts, which are oblivious
to linear order.

The input and output polarities of quantifiers are highly mutually constrained.
Take \phrase{everybody} for example.  If we hold the polarity assignments of the
other quantifiers fixed, then the existence of a linear-scope reading for
\phrase{A man introduced everybody to somebody} forces \phrase{everybody} to be
input-positive and output-neutral.  But then our account
predicts that \phrase{Nobody introduced everybody to somebody} has
a linear-scope reading, unlike the simpler sentence \phrase{Nobody introduced
Alice to somebody}.  This prediction is borne out, as observed by \citet[pages
121--122]{kroch-semantics} and discussed by \citet{szabolcsi-positive}.

\section*{Acknowledgments}

Thanks to Chris Barker, Raffaella Bernardi, William Ladusaw, Richard Moot, Chris
Potts, Stuart Shieber, Dylan Thurston, and three anonymous referees.  This work
is supported by the United States National Science Foundation Grant BCS-0236592.

\bibliographystyle{acl}
\bibliography{ccshan}

\end{document}